%% file: main.tex
\documentclass{article}
\usepackage{iclr2020_conference, times}
\iclrfinalcopy
\usepackage{natbib}

\usepackage{authblk}
\usepackage{hyperref}
\usepackage{comment}
\usepackage[noend]{algpseudocode}
\usepackage{caption}
\usepackage{subcaption}
\usepackage{algorithm}
\usepackage{footnote}
\usepackage{amssymb}%
\usepackage{pifont}%

\usepackage{xcolor}
\definecolor{darkgreen}{rgb}{0.0, 0.5, 0.0}
\definecolor{darkred}{rgb}{0.8, 0.0, 0.0}
\newcommand{\cmark}{\color{darkgreen}{\ding{51}}}%
\newcommand{\xmark}{\color{darkred}{\ding{55}}}%

\input{header}

\makeatletter
\def\blfootnote{\gdef\@thefnmark{*}\@footnotetext}
\makeatother

\title{Implementation Matters in Deep Policy\\Gradients: A Case Study on PPO and TRPO}

\author[1*]{Logan Engstrom}
\author[1*]{Andrew Ilyas}
\author[1]{Shibani Santurkar}

\author[1]{Dimitris Tsipras}
\author[2]{\\Firdaus Janoos}
\author[1,2]{Larry Rudolph}
\author[1]{Aleksander M\k{a}dry}

\affil[ ]{${}^{1}$MIT\ \ \ \ ${}^{2}$Two Sigma}
\affil[ ]{\texttt{\{engstrom,ailyas,shibani,tsipras,madry\}@mit.edu}}
\affil[ ]{\texttt{rudolph@csail.mit.edu, firdaus.janoos@twosigma.com}}

\begin{document}
\blfootnote{Equal contribution. Work done in part while interning at Two Sigma.}
\maketitle

\input{abstract}
\input{introduction}
\input{rel-work}
\input{case-study}
\input{trust_region}

\input{disentangling}
\input{conclusion}

\section{Acknowledgements}
We would like to thank Chloe Hsu for identifying a bug in our initial
implementation of PPO and TPRO. Work supported in part by the NSF grants
CCF-1553428, CNS-1815221, the Google PhD Fellowship, the Open Phil AI
Fellowship, and the Microsoft Corporation. 

\bibliography{paper.bib}
\bibliographystyle{iclr2020_conference}

\clearpage
\appendix

\input{appendix}

\end{document}

%% file: header.tex
\usepackage{amsmath}
\usepackage{booktabs}
\usepackage[T1]{fontenc}    %
\usepackage{amssymb}
\usepackage{url}
\usepackage{amsthm}
\usepackage{array}
\usepackage{tabularx}
\usepackage{subcaption}
\usepackage{caption}
\usepackage{multirow}
\usepackage{enumitem}
\usepackage{adjustbox}

\usepackage{color}
\usepackage{graphicx}

\newcolumntype{C}{>{\centering}X}%

\usepackage{thmtools}
\usepackage{thm-restate}

%% file: abstract.tex
\begin{abstract}
We study the roots of algorithmic progress in deep policy gradient algorithms
through a case study on two popular algorithms: Proximal Policy Optimization
(PPO) and Trust Region Policy Optimization (TRPO). Specifically, we investigate
the consequences of ``code-level optimizations:'' algorithm augmentations found
only in implementations or described as auxiliary details to the core algorithm.
Seemingly of secondary importance, such optimizations turn out to have a major
impact on agent behavior. Our results show that they (a) are responsible for
most of PPO's gain in cumulative reward over TRPO, and (b) fundamentally change
how RL methods function. These insights show the difficulty and importance of
attributing performance gains in deep reinforcement learning. 
\end{abstract}

%% file: introduction.tex
\section{Introduction}
\label{sec:intro}
Deep reinforcement learning (RL) algorithms have fueled many of the most
publicized achievements in modern machine
learning~\citep{silver2017mastering,dota,abbeel2016deep,mnih2013playing}.
However, despite these accomplishments, deep RL methods still are not nearly as
reliable as their (deep) supervised learning counterparts. Indeed, recent research found
the existing deep RL methods to be
brittle~\citep{henderson2017deep,zhang2018natural}, hard to
reproduce~\citep{henderson2017deep, Tucker2018TheMO}, unreliable across
runs~\citep{henderson2017deep,henderson2018did}, and sometimes outperformed by simple
baselines~\citep{Mania2018SimpleRS}.

The prevalence of these issues points to a broader problem: we do not understand
how the parts comprising deep RL algorithms impact agent training, either
separately or as a whole. This unsatisfactory understanding suggests that we
should re-evaluate the inner workings of our algorithms. Indeed, the overall
question motivating our work is: how do the multitude of mechanisms used in deep
RL training algorithms impact agent behavior?

\paragraph{Our contributions.} %
We analyze the underpinnings of agent behavior---both through the traditional metric of 
cumulative reward, and by measuring more fine-grained algorithmic properties. As a
first step, we conduct a case study of two of the
most popular deep policy-gradient methods: Trust Region Policy Optimization
(TRPO)~\citep{trpo} and Proximal Policy Optimization
(PPO)~\citep{schulman2017proximal}. These two methods are closely related: PPO
was originally developed as a refinement of TRPO. 

We find that much of the observed improvement in reward brought by PPO may come from
seemingly small modifications to the core algorithm which we call {\em code-level
optimizations}. These optimizations are either found only in implementations of
PPO, or are described as auxiliary details and are not present in the
corresponding TRPO baselines\footnote{Note that these code-level optimizations
are separate from ``implementation choices'' like the choice of PyTorch
versus TensorFlow in that they intentionally change the training algorithm's
operation.}. We pinpoint these modifications, and perform an ablation study
demonstrating that they are instrumental to the PPO's performance.

This observation prompts us to study how code-level optimizations change
agent training dynamics, and whether we can truly think of these optimizations as merely
auxiliary improvements. Our results indicate that these optimizations
fundamentally change algorithms' operation, and go even beyond improvements in agent
reward. We find that they majorly impact 
a key algorithmic principle behind TRPO and PPO's operations: trust region
enforcement. 

Ultimately, we discover that the \emph{PPO code-optimizations are more
important in terms of final reward achieved} than the choice of general training
algorithm (TRPO vs. PPO). This result is in stark contrast to the previous view
that the central PPO clipping method drives the gains seen in
\citet{schulman2017proximal}. In doing so, we demonstrate that the algorithmic
changes imposed by such optimizations make rigorous comparisons of algorithms
difficult. Without a rigorous understanding of the full impact of code-level
optimizations, we cannot hope to gain any reliable insight from comparing
algorithms on benchmark tasks.

Our results emphasize the importance of building RL methods in a modular manner.
To progress towards more performant and reliable algorithms, we need to
understand each component's impact on agents' behavior and performance---both
individually, and as part of a whole.

Code for all the results shown in this work is available at
\url{https://github.com/MadryLab/implementation-matters}.

%% file: rel-work.tex
\section{Related Work}
The idea of using gradient estimates to update neural network--based RL agents
dates back at least to the work of \citet{Williams1992SimpleSG}, who proposed
the REINFORCE algorithm. Later, \citet{Sutton1999PolicyGM} established a
unifying framework that casts the previous algorithms as instances of the policy
gradient method.

Our work focuses on proximal policy optimization
(PPO)~\citep{schulman2017proximal} and trust region policy optimization
(TRPO)~\citep{trpo}, which are two of the most prominent policy
gradient algorithms used in deep RL. Much of the original inspiration for the
usage of the trust regions stems from the conservative policy update of
\citet{Kakade2001ANP}. This policy update, similarly to TRPO, uses a natural
gradient descent-based greedy policy update. TRPO also bears similarity to the
relative policy entropy search method of \citet{Peters2010RelativeEP}, which
constrains the distance between marginal action distributions (whereas TRPO
constrains the conditionals of such action distributions).

Notably, \citet{henderson2017deep} points out a number of brittleness,
reproducibility, and experimental practice issues in deep RL algorithms.
Importantly, we build on the observation of \citet{henderson2017deep} that final
reward for a given algorithm is greatly influenced depending on the code base
used. \citet{Rajeswaran2017TowardsGA} and \citet{Mania2018SimpleRS} also
demonstrate that on many of the benchmark tasks, the performance of PPO and TRPO
can be matched by fairly elementary randomized search approaches. Additionally,
\citet{Tucker2018TheMO} showed that one of the recently proposed extensions of
the policy gradient framework, i.e., the usage of baseline functions that are
also action-dependent (in addition to being state-dependent), might not lead to
better policies after all.

%% file: case-study.tex
\section{Attributing Success in Proximal Policy Optimization}
\label{sec:case-study}
Our overarching goal is to better understand the underpinnings of the behavior
of deep policy gradient methods. We thus perform a careful study of two tightly
linked algorithms: TRPO and PPO (recall that PPO is motivated as TRPO with a
different trust region enforcement mechanism). To better understand these
methods, we start by thoroughly investigating their implementations in practice.
We find that in comparison to TRPO, the PPO implementation contains many non-trivial optimizations
that are not (or only barely) described in its corresponding paper. Indeed,
the standard implementation of PPO\footnote{From the OpenAI baselines GitHub
  repository: \url{https://github.com/openai/baselines}} contains the following
additional optimizations:
\begin{enumerate}
\item \textbf{Value function clipping:}
	\citet{schulman2017proximal} originally
	suggest fitting the value network via regression to target values:
	$$L^{V} = (V_{\theta_t} - V_{targ})^2,$$
	but the standard implementation instead fits the value network with a
	PPO-like objective:
	$$L^{V} = \max\left[\left(V_{\theta_t} - V_{targ}\right)^2,  
	  \left(\text{clip}\left(V_{\theta_t}, V_{\theta_{t-1}}-\varepsilon,
	  V_{\theta_{t-1}} + \varepsilon\right) - V_{targ}\right)^2\right],$$
	where $V_\theta$ is clipped around the previous
	value estimates (and $\varepsilon$ is fixed to the same value as
	  the value used to clip probability ratios in the PPO loss function (cf.
	  Eq.~\eqref{eqn:ppo} in Section~\ref{sec:algo_effects}).
\item \textbf{Reward scaling:} Rather than feeding the 
	rewards directly from the environment into the objective, the PPO
	implementation performs a certain discount-based scaling scheme.
	In this scheme, the rewards are divided through by the standard deviation of
	a rolling discounted sum of the rewards (without subtracting and
	re-adding the mean)---see Algorithm~\ref{alg:ppo-norm} in
	Appendix~\ref{app:impl_opt}. 
\item \textbf{Orthogonal initialization and layer scaling:} Instead of
	using the default weight initialization scheme for the policy and
	value networks, the implementation
	uses an orthogonal initialization scheme with scaling that varies
	from layer to layer.
\item \textbf{Adam learning rate annealing:} Depending on the task, the
	implementation sometimes anneals the learning rate of
	Adam~\citep{Kingma2014AdamAM} (an already adaptive method) for optimization.
\item \textbf{Reward Clipping}: The implementation also
  clips the rewards within a preset range (usually $[-5, 5]$ or $[-10, 10]$).
\item \textbf{Observation Normalization}: In a similar manner to the
  rewards, the raw states are also not fed into the optimizer. Instead, the
  states are first normalized to mean-zero, variance-one vectors.
\item \textbf{Observation Clipping}: Analagously to rewards, the
  observations are also clipped within a range, usually $[-10, 10]$.
\item \textbf{Hyperbolic tan activations}: As observed
  by~\cite{henderson2017deep}, implementations of policy gradient algorithms
  also use hyperbolic tangent function activations between layers in the
  policy and value networks.
\item \textbf{Global Gradient Clipping}: After computing the gradient with
  respect to the policy and the value networks, the implementation clips
  the gradients such the ``global $\ell_2$ norm'' (i.e. the norm of the
  concatenated gradients of all parameters) does not exceed $0.5$. 
\end{enumerate}

\begin{figure}[!ht]
  \begin{center}
      Humanoid-v2 \\
	\includegraphics[width=\textwidth]{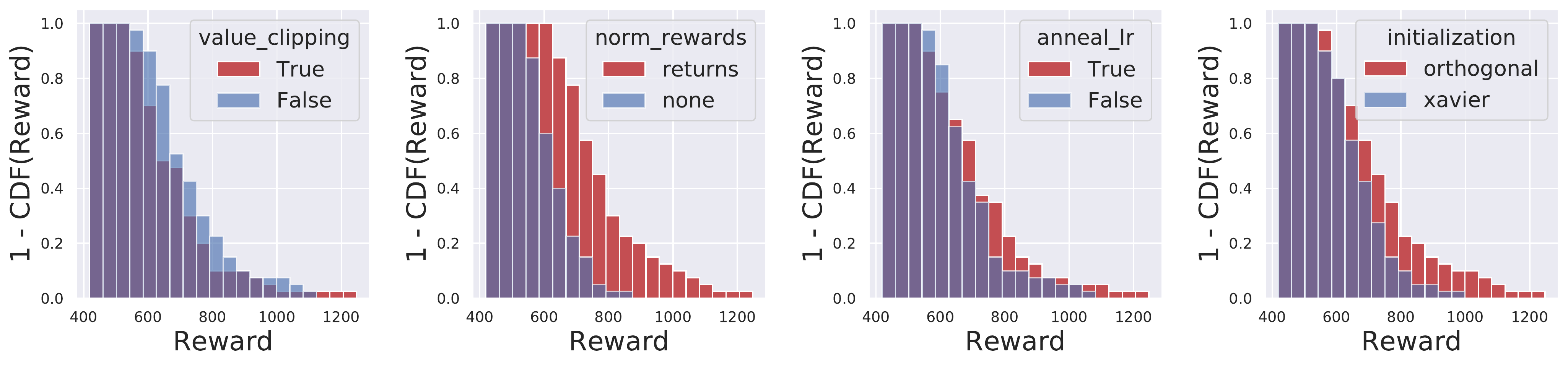}
      Walker2d-v2 \\
	\includegraphics[width=\textwidth]{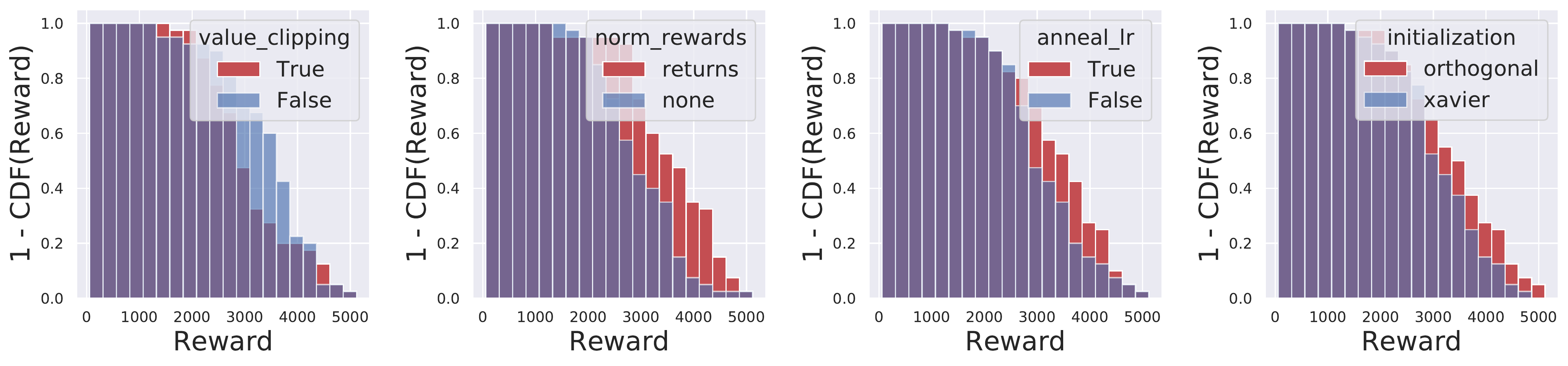}
\end{center}
\caption{An ablation study on the first four optimizations described in
  Section~\ref{sec:case-study} (value clipping, reward scaling, network
  initialization, and learning rate annealing). For each of the $2^4$ possible
  configurations of optimizations, we train a Humanoid-v2 (top) and
  Walker2d-v2 (bottom) agent using PPO with
  five random seeds and a grid of learning rates, and choose the learning rate
  which gives the best average reward (averaged over the random seeds). We then
  consider all rewards from the ``best learning rate'' runs (a total of $5\times
  2^4$ agents), and plot histograms in which agents are partitioned based on
  whether each optimization is on or off. Our results show that reward
  normalization, Adam annealing, and network initialization each significantly
  impact the rewards landscape with respect to hyperparameters, and were
  necessary for attaining the highest PPO reward within the tested
  hyperparameter grid. We detail our experimental setup in
  Appendix~\ref{app:exp_setup}.} 
\label{fig:ablation-histogram}
\end{figure}

These optimizations may appear as merely surface-level or insignificant
algorithmic changes to the core policy gradient method at hand. However, we find
that they dramatically impact the performance of PPO. Specifically, we
perform a full ablation study on the four optimizations mentioned
above\footnote{Due to restrictions on computational resources, we could only
perform a full ablation on the first four of the identified optimizations.}.
Figure~\ref{fig:ablation-histogram} shows a
histogram of the final rewards of agents trained with every possible
configuration of the above optimizations---for each configuration, a grid search
for the optimal learning rate is performed, and we measure the reward of random
agents trained using the identified learning rate. Our findings suggest that
many code-level optimizations are necessary for PPO to attain its claimed performance.

The above findings show that our ability to understand PPO from an algorithmic
perspective hinges on the ability to distill out its fundamental principles from
such algorithm-independent (in the sense that these optimizations can be
implemented for any policy gradient method) optimizations. We thus consider a
variant of PPO called \textsc{PPO-Minimal} (PPO-M) which implements only the
core of the algorithm. PPO-M uses the standard value network loss, no reward
scaling, the default network initialization, and Adam with a fixed learning
rate. Importantly, PPO-M ignores all the code-level optimizations listed 
at the beginning of Section~\ref{sec:case-study}. We explore PPO-M
alongside PPO and TRPO. We list all the algorithms we study and their defining
properties in Table~\ref{tab:algorithms}.

\begin{table}[ht]
\centering
\caption{An overview of the algorithms studied in this work.
 Step method refers to the method used to build each training step,
  \textsc{PPO} clipping refers to the use of clipping in the step (as in
  Equation~\eqref{eqn:ppo}), and \textsc{PPO} optimizations refer to the
  optimizations listed in Section~\ref{sec:case-study}.}
\begin{tabular}{@{}lcccc@{}}
\toprule
\textbf{Algorithm}  &\textbf{Section}  & \textbf{Step method}  & \textbf{Uses \textsc{PPO} clipping?} & \textbf{Uses \textsc{PPO} optimizations?} \\
\midrule
\textsc{PPO}   & ---     & \textsc{PPO}  & \cmark & As in~\citep{baselines}  \\
\textsc{PPO-M} & Sec. \ref{sec:case-study}     & \textsc{PPO}    & \cmark & \xmark \\
\textsc{PPO-NoClip} & Sec. \ref{sec:algo_effects} & \textsc{PPO}   & \xmark & Found via grid search \\
\textsc{TRPO}   & ---    & \textsc{TRPO}  & --- & \xmark \\
\textsc{TRPO+}  & Sec. \ref{sec:disentangling} & \textsc{TRPO}  & --- & Found via grid search \\
\bottomrule
\end{tabular}
\label{tab:algorithms}
\end{table}

Overall, our results on the importance of these optimizations both corroborate
results demonstrating the brittleness of deep policy gradient methods, and
demonstrate that even beyond environmental brittleness, the algorithms
themselves exhibit high sensitivity to implementation choices~\footnote{This
  might also explain the difference between different codebases observed
  in~\cite{henderson2017deep}}.

%% file: trust_region.tex
\section{Code-Level Optimizations have Algorithmic Effects} 
\label{sec:algo_effects}

The seemingly disproportionate effect of code-level optimizations identified in
our ablation study may lead us to ask: how do these seemingly superficial
code-level optimizations impact underlying agent behavior? In this section,
we demonstrate that the code-level optimizations fundamentally alter agent
behavior. Rather than merely improving ultimate cumulative award, such
optimizations directly impact the principles motivating the core algorithms.

\paragraph{Trust Region Optimization.}
A key property of policy gradient algorithms is that update steps computed at
any specific policy $\pi_{\theta_t}$ are only guaranteed predictiveness in a
neighborhood around $\theta_t$. Thus, to ensure that the update steps we derive
remain predictive, many policy gradient algorithms ensure that these steps stay
in the vicinity of the current policy. The resulting ``trust region''
methods~\citep{Kakade2001ANP, trpo, schulman2017proximal} try to constrain the
local variation of the parameters in policy-space by restricting the
distributional distance between successive policies.

A popular method in this class is trust region policy optimization
(TRPO)~\cite{trpo}. TRPO constrains the KL divergence
between successive policies on the optimization trajectory, leading to the
following problem:
\begin{align}
\max_\theta\quad &\mathbb{E}_{(s_t, a_t) \sim
\pi}\left[\frac{\pi_\theta(a_t|s_t)}{\pi(a_t|s_t)}\widehat{A}_{\pi}(s_t, a_t)\right] \nonumber \\
\text{s.t.}\quad &D_{KL}(\pi_\theta(\cdot \mid s)||\pi(\cdot\mid s)) \leq \delta,\quad \forall s\;. \label{eqn:trpo}
\end{align}
In practice, we maximize this objective with a second-order approximation
of the KL divergence and natural gradient descent, and replace the
worst-case KL constraints over all possible states with an approximation of
the mean KL based on the states observed in the current trajectory.

\paragraph{Proximal policy optimization.}  
One disadvantage of the TRPO algorithm is that it can be
computationally costly---the step direction is
estimated with nonlinear conjugate gradients, which requires the
computation of multiple Hessian-vector products. To address this
issue, \citet{schulman2017proximal} propose proximal policy
optimization (PPO), which tries to enforce a trust region with a different
objective that does not require computing a projection. Concretely, PPO
proposes replacing the KL-constrained objective~\eqref{eqn:trpo} of TRPO
by clipping the objective function directly as:
\begin{align}
&\max_\theta\ \mathbb{E}_{(s_t, a_t) \sim
\pi}\left[\min\left(\text{clip}\left(\rho_t, 1-\varepsilon,
1+\varepsilon\right)\widehat{A}_{\pi}(s_t, a_t),\ 
\rho_t\widehat{A}_{\pi}(s_t, a_t)\right)\right] \label{eqn:ppo}
\end{align}
where
\begin{align}
\rho_t = \frac{\pi_\theta(a_t|s_t)}{\pi(a_t|s_t)}.
\end{align}
Note that this objective can be optimized without an explicit projection step,
leading to a simpler parameter update during training. In addition to its
simplicity, PPO is intended to be faster and more sample-efficient than
TRPO~\citep{schulman2017proximal}. 

\paragraph{Trust regions in TRPO and PPO.} Enforcing a trust
region is a core algorithmic property of different policy gradient
methods. However, whether or not a trust region is enforced is not directly
observable from the final rewards. So, how does this algorithmic property vary
across state-of-the-art policy gradient methods?

\begin{figure}[tb]
    \begin{subfigure}[b]{1\textwidth}
	    \includegraphics[width=1\textwidth]{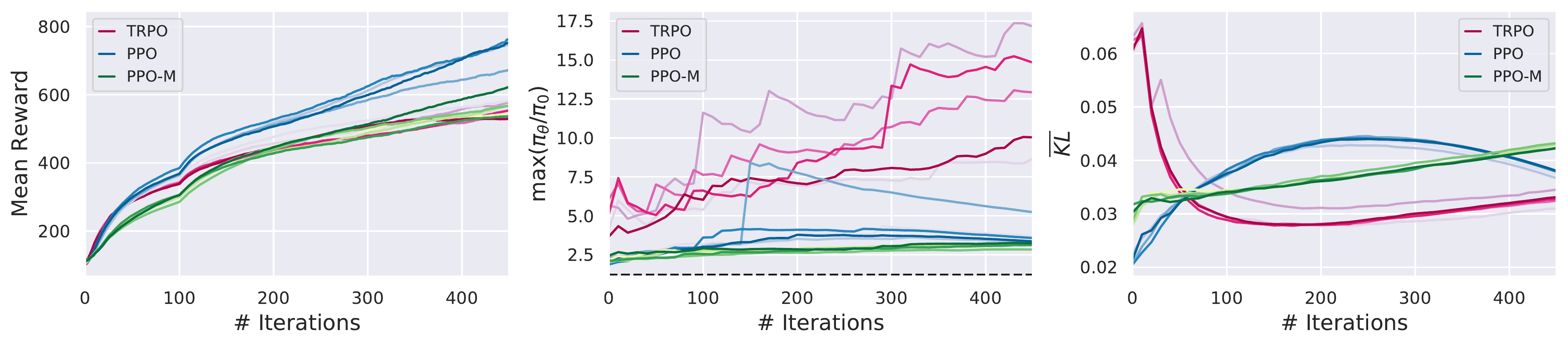} 
    \end{subfigure}
    \caption{Per step mean reward, maximum ratio (c.f. \eqref{eqn:ppo}), mean
      KL, and mean KL for agents trained to solve the MuJoCo Humanoid-v2 task.
      The quantities are measured over the state-action pairs collected in the
      \textit{training step}. Each line represents a training curve from a
      separate agent. The black dotted line represents the $1+\epsilon$ ratio
      constraint in the PPO algorithm, and we measure each quantity every twenty
      five steps. We take mean and max over the KL divergences between the
      conditional action distributions of successive policies. In the left plot,
      the reward for each trained agent. In the middle plot, the PPO variants'
      maximum ratios consistently violate the ratio 
      ``trust region.'' In the right plot, both PPO and PPO-M
      constraint the KL well (compared to the TRPO bound of 0.07). The two
      methods exhibit different behavior.
      We measure the quantities over a \textit{heldout} set
      of state-action pairs and find little qualitative difference in the
      results (seen in Figure~\ref{app:trust_regions_heldout} in the appendix),
      suggesting that TRPO enforces a mean KL trust region. We show
      plots for additional tasks in the Appendix in
      Figure~\ref{app:trust_regions_train}. We detail our experimental setup in
      Appendix~\ref{app:exp_setup}.}
    \label{fig:trust_region}
\end{figure}

In Figure~\ref{fig:trust_region} we measure the mean KL divergence between
successive policies in a training run of both TRPO and PPO-M (PPO without
code-level optimizations). Recall that TRPO is designed specifically to
constrain this KL-based trust region, while the clipping mechanism of PPO
attempts to approximate it. Indeed, we find that TRPO precisely
enforces this trust region (this is unsuprising, and nearly by construction).

We thus turn our attention to the trust regions induced by training with PPO and
PPO-M. 
First, we consider mathematically the contribution of a single state-action pair
to the gradient of the PPO objective, which is given by
\[
    \nabla_\theta L_{PPO} = \begin{cases}
	\nabla_\theta L_\theta
	& \frac{\pi_\theta(a|s)}{\pi(a|s)} \in
	[1-\epsilon, 1+\epsilon] \text{ or } L^{C}_\theta < L_\theta \\
	0 & \text{otherwise}
    \end{cases},
\]
\[
    \text{where} \qquad 
    L_\theta :=  \mathbb{E}_{(s, a) \in \tau \sim
	    \pi}\left[\frac{\pi_\theta(a|s)}{\pi(a|s)}
	A_{\pi}(s, a)\right],
	\] \[
    \text{and} \qquad 
	L^{C}_\theta :=  \mathbb{E}_{(s, a) \in \tau \sim
	    \pi}\left[
	    \text{clip}\left(\frac{\pi_\theta(a|s)}{\pi(a|s)},
	    1-\varepsilon, 1+\varepsilon\right)
	A_{\pi}(s, a)\right]
\]
are respectively the standard and clipped versions of the surrogate objective.
As a result, since we initialize $\pi_\theta$ as $\pi$ (and thus the
ratios start all equal to one) the first step we take is identical to a
maximization step over the \textit{unclipped} surrogate objective. It thus
stands to reason that the nature of the trust region enforced is heavily
dependent on the {\em method} with which the clipped PPO objective is
optimized, rather than on the objective itself. Therefore, the size of the step
we take is determined solely by the steepness of the surrogate landscape
(i.e. Lipschitz constant of the optimization problem we solve), and we can
end up moving arbitrarily far from the trust region. We hypothesize that
this dependence of PPO on properties of the optimizer rather than on the
optimization objective contributes to the brittleness of the algorithm to
hyperparameters such as learning rate and momentum, as observed
by~\citet{henderson2018did} and others.

The results we observe (shown in Figure~\ref{fig:trust_region}) corroborate this
intuition. For agents trained with optimal parameters, all three algorithms are
able to maintain a KL-based trust region. First, we note that all three
algorithms fail to maintain a ratio-based trust region, despite PPO and PPO-M
being trained directly with a ratio-clipping objective. Furthermore, the nature
of the KL trust region enforced differs between PPO and PPO-M, despite the fact
that the core algorithm remains constant between the two methods; while PPO-M KL
trends up as the number of iterations increases, PPO KL peaks halfway through
training before trending down again.

The findings from this experiment and the corresponding calculations demonstrate
that perhaps a key factor in the behavior of PPO-trained agents even from an
algorithmic viewpoint comes from auxiliary optimizations, rather than the
core methodology.

%% file: disentangling.tex
\section{Identifying Roots of Algorithmic Progress}
\label{sec:disentangling}
State-of-the-art deep policy gradient methods are comprised of many interacting
components. At what is generally described as their core, these methods
incorporate mechanisms like trust region-enforcing steps, time-dependent value
predictors, and advantage estimation methods for controlling the
exploitation/exploration trade-off~\citep{schulman2015high}. However, these
algorithms also incorporate many less oft-discussed optimizations (cf.
Section~\ref{sec:case-study}) that ultimately dictate much of agent behavior
(cf. Section~\ref{sec:algo_effects}). Given the need to improve on these
algorithms, the fact that such optimizations are so important begs
the question: \textit{how do we identify the true roots of algorithmic progress
  in deep policy gradient methods?}

Unfortunately, answering this question is not easy. 
Going back to our study of PPO and TRPO, it is widely believed (and
claimed) that the key innovation of PPO responsible for its improved performance
over the baseline of TRPO is the ratio clipping mechanism discussed in 
Section~\ref{sec:algo_effects}. However, we have already shown that this
clipping mechanism is insufficient theoretically to maintain a trust region, and
also that the {\em method} by which the objective is optimized appears to have
significant effect on the resulting trust region. If code-level optimizations
are thus partially responsible for algorithmic properties of PPO, is
it possible that they are also a key factor in PPO's improved performance? 

To address this question, we set out to further disentangle the impact of PPO's core
clipping mechanism from its code-level optimizations by once again considering
variations on the PPO and TRPO algorithms. Specifically, we examine
how employing the core PPO and TRPO steps changes model performance while
controlling for the effect of code-level optimizations identified in standard
implementations of PPO (in particular, we focus on those covered in
Section~\ref{sec:case-study}). These
code-level optimizations are largely algorithm-independent, and so they can be
straightforwardly applied or lightly adapted to any policy gradient method. The
previously introduced PPO-M algorithm corresponds to PPO {\em
without} these optimizations. To further account for their effects, we study
an additional algorithm which we denote as {\em TRPO+}, consisting of the
core algorithmic contribution of TRPO in combination with PPO's code-level
optimizations as identified in Section~\ref{sec:case-study}~\footnote{We also
add a new code-level optimization, a {\em KL decay}, inapplicable to PPO but
meant to serve as the analog of Adam learning rate annealing.}. We note that
TRPO+ together with the other three algorithms introduced (PPO, PPO-M, and TRPO;
all listed in Table~\ref{tab:algorithms}) now capture all combinations of
core algorithms and code-level optimizations, allowing us to study the impact
of each in a fine-grained manner.

As our results show in Table~\ref{tab:trpoplus}, it turns out that code-level
optimizations contribute to algorithms' increased performance often
significantly more than the choice of algorithm (i.e., using PPO vs.
TRPO). For example, on Hopper-v2, PPO and TRPO see 17\% and 21\% improvements
(respectively) when equipped with code-level optimizations. At the same time,
for all tasks after fixing the choice to use or not use optimizations, the
core algorithm employed does not seem to have a significant impact on reward. In
Table~\ref{tab:trpoplus}, we quantify this contrast through the following
two metrics, which we denote {\em average algorithmic improvement}
(\textsc{AAI}) and {\em average code-level improvement} (\textsc{ACLI}):
$$\textsc{AAI} = \max\{|\textsc{PPO} - \textsc{TRPO+}|, |\textsc{PPO-M} - \textsc{TRPO}|\},$$
$$\textsc{ACLI} = \max\{|\textsc{PPO} - \textsc{PPO-M}|, |\textsc{TRPO+} -
\textsc{TRPO}|\}.$$
In short, AAI measures the maximal effect of switching step algorithms, whereas
ACLI measures the maximal effect of adding in code-level optimizations for a
fixed choice of step algorithm.  

\begin{table}[t!]
\centering
\caption{Full ablation of step choices (PPO or TRPO) and presence of code-level
  optimizations measuring agent performance on benchmark tasks. TRPO+ is a
  variant of TRPO that uses PPO inspired code-level optimizations, and PPO-M is
  a variant of PPO that does not use PPO's code-level optimizations (cf.
  Section~\ref{sec:case-study}). Varying the use of
  code-level optimizations impacts performance significantly more than varying
  whether the PPO or TRPO step is used. We detail our experimental setup
  in Appendix~\ref{app:exp_setup}. We train at least 80 agents for each
  estimate (more for some high-variance cases). We
  present 95\% confidence intervals computed via a 1000-sample bootstrap.
  We also present the AAI and ACLI metrics discussed in
  Section~\ref{sec:disentangling}, which attempt to quantify the relative
  contribution of algorithmic choice vs. use of code-level optimizations
  respectively.}
  {\begin{sc}
\begin{tabular}{@{}lccc@{}}
  \toprule
 &  \multicolumn{3}{c}{MuJoCo Task} \\
Step &              Walker2d-v2 &                Hopper-v2 &             Humanoid-v2 \\
\midrule
PPO   &  3292 [3157, 3426] &  2513 [2391, 2632] &    806 [785, 827] \\
PPO-M &  2735 [2602, 2866] &  2142 [2008, 2279] &    674 [656, 695] \\
TRPO  &  2791 [2709, 2873] &  2043 [1948, 2136] &    586 [576, 596] \\
TRPO+ &  3050 [2976, 3126] &  2466 [2381, 2549] &  1030 [979, 1083] \\ \midrule
AAI   &  242               &  99                &   224             \\
ACLI  &  557               &  421               &   444             \\
\bottomrule
\end{tabular}
  \end{sc}
}
\label{tab:trpoplus}
\end{table}

\paragraph{PPO without clipping.} Given the relative insignificance of the
step mechanism compared to the use of code-level optimizations, we are
prompted to ask: {\em to what extent is the clipping mechanism of PPO
actually responsible for the algorithm's success?} In
Table~\ref{tab:mundane_step}, we assess this by considering a
\textsc{PPO-NoClip} algorithm which makes use of common code-level
optimizations (by gridding over the best possible combination of such
optimizations) but does not employ a clipping mechanism (this is the same
algorithm we studied in Section~\ref{sec:algo_effects} in the context of
trust region enforcement)---recall that we list all the algorithms studied in
Table~\ref{tab:algorithms}.

It turns out that the clipping mechanism is not necessary to achieve
high performance---we find that \textsc{PPO-NoClip} performs uniformly better
than \textsc{PPO-M}, despite the latter employing the core PPO clipping
mechanism. Moreover, introducing code-level optimizations seems to outweigh even
the core PPO algorithm in terms of effect on rewards. In fact, we find that with sufficient hyperparameter tuning,
\textsc{PPO-NoClip} often matches the performance of standard \text{PPO},
which {\em includes} a standard configuration of code-level
optimizations\footnote{Note that it is possible that further refinement on the
code-level optimizations could be added on top of PPO to perhaps improve its
performance to an even greater extent (after all, \textsc{PPO-NoClip} can
only express a subset the training algorithms covered by PPO, as the latter
leaves the clipping severity $\varepsilon$ to be free parameter)}. We also
include benchmark PPO numbers from the OpenAI \texttt{baselines}
repository~\citep{baselines} (where available) to put results into context.

\begin{table}[t!]
\centering
\caption{Comparison of PPO performance to PPO without clipping. We find that
  there is little difference between the rewards attained by the two
  algorithms on the benchmark tasks. Note that both algorithms use code-level
  optimizations; our results indicate that the clipping mechanism is often of
  comparable or lesser importance to the use of code-level optimizations. We detail our
  experimental setup in Appendix~\ref{app:exp_setup}. We train at least 80
  agents for each estimate (for some high-variance cases, more agents were
  used). We present bootstrapped 95\% confidence intervals computed with 1000
  samples. We also present results from the OpenAI 
  baselines~\citep{baselines} where available.} {\begin{sc}
\begin{tabular}{@{}lccc@{}}
\toprule
{} &        Walker2d-v2 &          Hopper-v2 &     Humanoid-v2 \\
\midrule
PPO        &  3292 [3157, 3426] &  2513 [2391, 2632] &  806 [785, 827] \\
PPO (\texttt{baselines})  & 3424 &  2316 & --- \\
PPO-M      &  2735 [2602, 2866] &  2142 [2008, 2279] &  674 [656, 695] \\
PPO-NoClip &  2867 [2701, 3024] &  2371 [2316, 2424] &  831 [798, 869] \\
\bottomrule
\end{tabular}
  \end{sc}
}
\label{tab:mundane_step}
\end{table}

Our results suggest that it is difficult to attribute success to different
aspects of policy gradient algorithms without careful analysis.

%% file: conclusion.tex
\section{Conclusion}
In this work, we take a first step in examining how the mechanisms powering deep
policy gradient methods impact agents both in terms of achieved reward and
underlying algorithmic behavior. Wanting to understand agent operation from the
ground up, we take a deep dive into the operation of two of the most popular deep
policy gradient methods: TRPO and PPO. In doing so, we identify a number of
``code-level optimizations''---algorithm augmentations found only in algorithms'
implementations or described as auxiliary details in their presentation---and
find that these optimizations have a drastic effect on agent performance.

In fact, these seemingly unimportant optimizations fundamentally change
algorithm operation in ways unpredicted by the conceptual policy gradient
framework. Indeed, the optimizations 
often dictate the nature of the trust region enforced by policy gradient
algorithms, even controlling for the surrogate objective being optimized.
We go on to test the importance of code-level optimizations in agent
performance, and find that PPO's marked improvement over TRPO (and even
stochastic gradient descent) can be largely attributed to these optimizations.

Overall, our results highlight the necessity of designing deep RL methods in a
\textit{modular} manner. When building algorithms, we should understand
precisely how each component impacts agent training---both in terms of overall
performance and underlying algorithmic behavior. It is impossible to properly
attribute successes and failures in the complicated systems that make up deep RL
methods without such diligence. More broadly, our findings suggest that
developing an RL toolkit will require moving beyond the current benchmark-driven
evaluation model to a more fine-grained understanding of deep RL methods.

%% file: appendix.tex
\section{Appendix}
\label{sec:appendix}

\subsection{Experimental Setup}
\label{app:exp_setup}
All the hyperparameters used in this paper were obtained through grid
searches. For PPO the exact code-level optimizations and their
associated hyperparameters (e.g. coefficients for entropy regularization,
reward clipping, etc.) were taken from the OpenAI baselines
repository~\footnote{\url{https://github.com/openai/baselines}}, and
gridding is performed over the value function learning rate, the clipping
constant, and the learning rate schedule. In TRPO, we grid over the same
parameters (replacing learning rate schedule with the KL constraint), but
omit the code-level optimizations. For PPO-NoClip, we grid over the same
parameters as PPO, in addition to the configuration of code-level
optimizations (since we lack a good reference for what the optimal
configuration of these optimizations is). For TRPO+ we also grid over the
code-level optimizations, and also implement a ``KL schedule'' whereby the
KL constraint can change over training (analogous to the learning rate
annealing optimization in PPO). Finally, for PPO-M, we grid over the same
parameters as PPO (just learning rate schedules), without any code-level
optimizations. The final parameters for each algorithm are given below, and a
more detailed account is available in our code release: \url{https://github.com/MadryLab/implementation-matters}.

\begin{table}[ht]
\centering
\caption{Hyperparameters for all algorithms for Walker2d-v2.}
\label{app-tab:hyperparameters_walker}
\begin{adjustbox}{width=\textwidth,center}
\begin{tabular}{llllll}
	\toprule
	{} &            PPO &      TRPO & PPO-NoClip &     PPO-M &          TRPO+ \\
	\midrule
	Timesteps per iteration         &           2048 &      2048 &       2048 &      2048 &           2048 \\
	Discount factor ($\gamma$)        &           0.99 &      0.99 &       0.99 &      0.99 &           0.99 \\
	GAE discount ($\lambda$)          &           0.95 &      0.95 &       0.85 &      0.95 &           0.95 \\
	Value network LR                &         0.0003 &    0.0003 &     0.0006 &    0.0002 &         0.0001 \\
	Value network num. epochs       &             10 &        10 &         10 &        10 &             10 \\
	Policy network hidden layers    &       [64, 64] &  [64, 64] &   [64, 64] &  [64, 64] &       [64, 64] \\
	Value network hidden layers     &       [64, 64] &  [64, 64] &   [64, 64] &  [64, 64] &       [64, 64] \\
	KL constraint ($\delta$)         &            N/A &      0.04 &        N/A &       N/A &           0.07 \\
	Fisher estimation fraction      &            N/A &       0.1 &        N/A &       N/A &            0.1 \\
	Conjugate gradient steps        &            N/A &        10 &        N/A &       N/A &             10 \\
	Conjugate gradient damping      &            N/A &       0.1 &        N/A &       N/A &            0.1 \\
	Backtracking steps              &            N/A &        10 &        N/A &       N/A &             10 \\
	Policy LR (Adam)                &         0.0004 &       N/A &   7.25e-05 &    0.0001 &            N/A \\
	Policy epochs                   &             10 &       N/A &         10 &        10 &            N/A \\
	PPO Clipping $\varepsilon$       &            0.2 &       N/A &      1e+32 &       0.2 &            N/A \\
	Entropy coeff.                  &              0 &         0 &      -0.01 &         0 &              0 \\
	Reward clipping                 &  [-10.0, 10.0] &        -- &  [-30, 30] &        -- &  [-10.0, 10.0] \\
	Gradient clipping ($\ell_2$ norm) &             -1 &        -1 &        0.1 &        -1 &              1 \\
	Reward normalization            &        returns &      none &    rewards &      none &        returns \\
	State clipping                  &  [-10.0, 10.0] &        -- &  [-30, 30] &        -- &  [-10.0, 10.0] \\
	\bottomrule
\end{tabular}
\end{adjustbox}
\end{table}

\begin{table}[ht]
\centering
\caption{Hyperparameters for all algorithms for Humanoid-v2.}
\label{app-tab:hyperparameters_humanoid}
\begin{adjustbox}{width=\textwidth,center}
\begin{tabular}{llllll}
	\toprule
	{} &            PPO &      TRPO &     PPO-NoClip &     PPO-M &          TRPO+ \\
	\midrule
	Timesteps per iteration         &           2048 &      2048 &           2048 &      2048 &           2048 \\
	Discount factor ($\gamma$)        &           0.99 &      0.99 &           0.99 &      0.99 &           0.99 \\
	GAE discount ($\lambda$)          &           0.95 &      0.95 &           0.85 &      0.95 &           0.85 \\
	Value network LR                &         0.0001 &    0.0003 &          5e-05 &    0.0004 &          5e-05 \\
	Value network num. epochs       &             10 &        10 &             10 &        10 &             10 \\
	Policy network hidden layers    &       [64, 64] &  [64, 64] &       [64, 64] &  [64, 64] &       [64, 64] \\
	Value network hidden layers     &       [64, 64] &  [64, 64] &       [64, 64] &  [64, 64] &       [64, 64] \\
	KL constraint ($\delta$)         &            N/A &      0.07 &            N/A &       N/A &            0.1 \\
	Fisher estimation fraction      &            N/A &       0.1 &            N/A &       N/A &            0.1 \\
	Conjugate gradient steps        &            N/A &        10 &            N/A &       N/A &             10 \\
	Conjugate gradient damping      &            N/A &       0.1 &            N/A &       N/A &            0.1 \\
	Backtracking steps              &            N/A &        10 &            N/A &       N/A &             10 \\
	Policy LR (Adam)                &        0.00015 &       N/A &          2e-05 &     9e-05 &            N/A \\
	Policy epochs                   &             10 &       N/A &             10 &        10 &            N/A \\
	PPO Clipping $\varepsilon$       &            0.2 &       N/A &          1e+32 &       0.2 &            N/A \\
	Entropy coeff.                  &              0 &         0 &           0.005 &         0 &              0 \\
	Reward clipping                 &  [-10.0, 10.0] &        -- &  [-10.0, 10.0] &        -- &  [-10.0, 10.0] \\
	Gradient clipping ($\ell_2$ norm) &             -1 &        -1 &            0.5 &        -1 &            0.5 \\
	Reward normalization            &        returns &      none &        returns &      none &        returns \\
	State clipping                  &  [-10.0, 10.0] &        -- &  [-10.0, 10.0] &        -- &  [-10.0, 10.0] \\
	\bottomrule
	\end{tabular}
\end{adjustbox}
\end{table}

\begin{table}[ht]
\centering
\caption{Hyperparameters for all algorithms for Hopper-v2.}
\label{app-tab:hyperparameters_hopper}
\begin{adjustbox}{width=\textwidth,center}
\begin{tabular}{llllll}
	\toprule
	{} &            PPO &      TRPO &   PPO-NoClip &     PPO-M &          TRPO+ \\
	\midrule
	Timesteps per iteration         &           2048 &      2048 &         2048 &      2048 &           2048 \\
	Discount factor ($\gamma$)        &           0.99 &      0.99 &         0.99 &      0.99 &           0.99 \\
	GAE discount ($\lambda$)          &           0.95 &      0.95 &        0.925 &      0.95 &           0.95 \\
	Value network LR                &        0.00025 &    0.0002 &       0.0004 &    0.0004 &         0.0002 \\
	Value network num. epochs       &             10 &        10 &           10 &        10 &             10 \\
	Policy network hidden layers    &       [64, 64] &  [64, 64] &     [64, 64] &  [64, 64] &       [64, 64] \\
	Value network hidden layers     &       [64, 64] &  [64, 64] &     [64, 64] &  [64, 64] &       [64, 64] \\
	KL constraint ($\delta$)         &            N/A &      0.13 &          N/A &       N/A &           0.04 \\
	Fisher estimation fraction      &            N/A &       0.1 &          N/A &       N/A &            0.1 \\
	Conjugate gradient steps        &            N/A &        10 &          N/A &       N/A &             10 \\
	Conjugate gradient damping      &            N/A &       0.1 &          N/A &       N/A &            0.1 \\
	Backtracking steps              &            N/A &        10 &          N/A &       N/A &             10 \\
	Policy LR (Adam)                &         0.0003 &       N/A &        6e-05 &   0.00017 &            N/A \\
	Policy epochs                   &             10 &       N/A &           10 &        10 &            N/A \\
	PPO Clipping $\varepsilon$       &            0.2 &       N/A &        1e+32 &       0.2 &            N/A \\
	Entropy coeff.                  &              0 &         0 &        -0.005 &         0 &              0 \\
	Reward clipping                 &  [-10.0, 10.0] &        -- &  [-2.5, 2.5] &        -- &  [-10.0, 10.0] \\
	Gradient clipping ($\ell_2$ norm) &             -1 &        -1 &            4 &        -1 &              1 \\
	Reward normalization            &        returns &      none &      rewards &      none &        returns \\
	State clipping                  &  [-10.0, 10.0] &        -- &  [-2.5, 2.5] &        -- &  [-10.0, 10.0] \\
	\bottomrule
	\end{tabular}
\end{adjustbox}
\end{table}

All error bars we plot are 95\% confidence intervals, obtained via
bootstrapped sampling.

\clearpage
\subsection{PPO Code-Level Optimizations}
\label{app:impl_opt}

\input{Figures/alg/norm_alg.tex}

\clearpage
\subsection{Trust Region Optimization}
\begin{figure}[!ht]
	\begin{subfigure}[b]{1\textwidth}
		\includegraphics[width=1\textwidth]{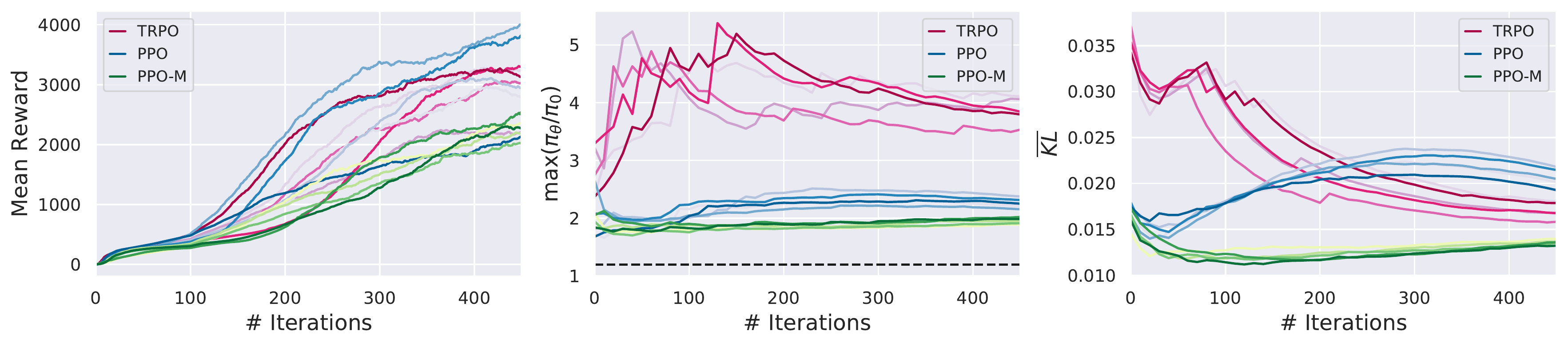} 
		\caption{Walker2d-v2 (train)}
	\end{subfigure}
	\begin{subfigure}[b]{1\textwidth}
		\includegraphics[width=1\textwidth]{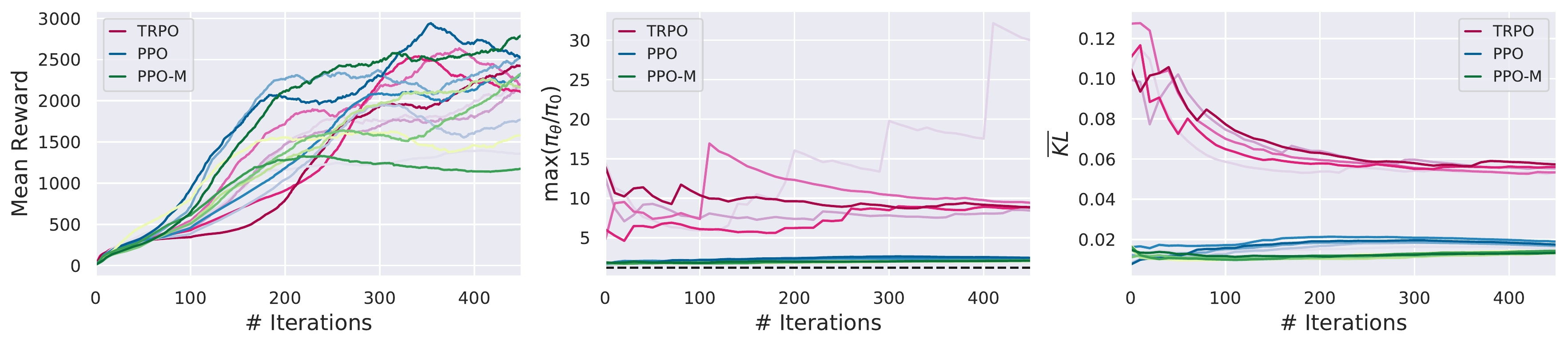} 
		\caption{Hopper-v2 (train)}
	\end{subfigure}
	
	\caption{Per step mean reward, maximum ratio (c.f. \eqref{eqn:ppo}), mean KL,
		and maximum versus mean
		KL for agents trained to solve the MuJoCo Humanoid task. The quantities are
		measured over the state-action pairs collected in the \textit{training step}.
		Each line represents a training curve from a
		separate agent.  The black dotted line
		represents the $1+\epsilon$ ratio constraint in the PPO algorithm, and we measure each
		quantity every twenty five steps. Compare the results here with
		Figure~\ref{app:trust_regions_heldout}; they are qualitatively nearly identical.}
	\label{app:trust_regions_train}
\end{figure}

\begin{figure}[!ht]
	\begin{subfigure}[b]{1\textwidth}
		\includegraphics[width=1\textwidth]{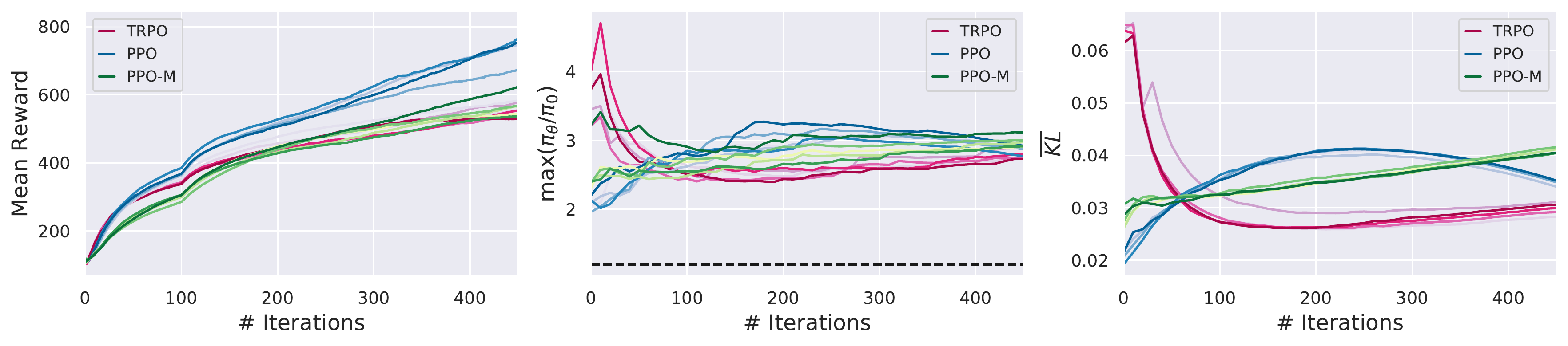} 
		\caption{Humanoid-v2 (heldout)}
	\end{subfigure}
	\begin{subfigure}[b]{1\textwidth}
		\includegraphics[width=1\textwidth]{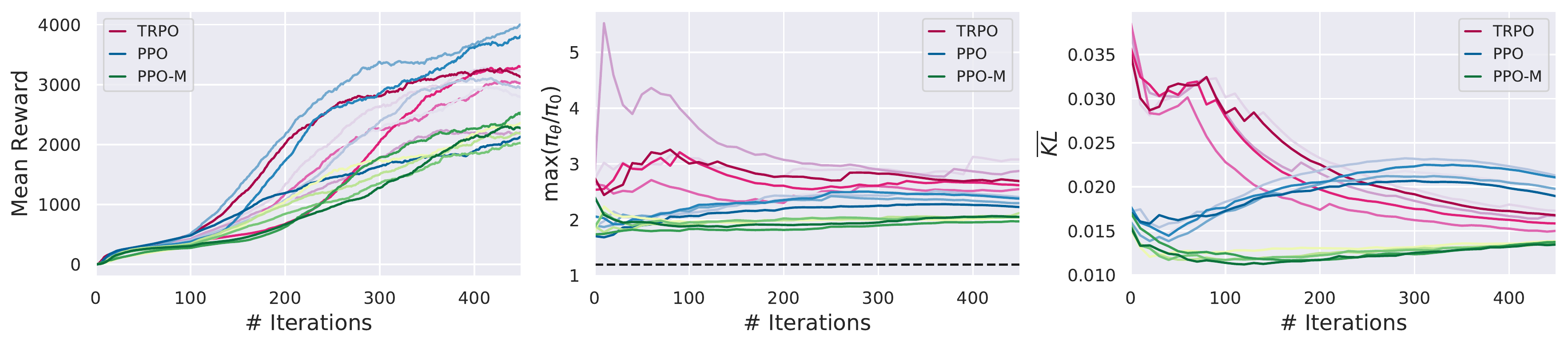} 
		\caption{Walker2d-v2 (heldout)}
	\end{subfigure}
	\begin{subfigure}[b]{1\textwidth}
		\includegraphics[width=1\textwidth]{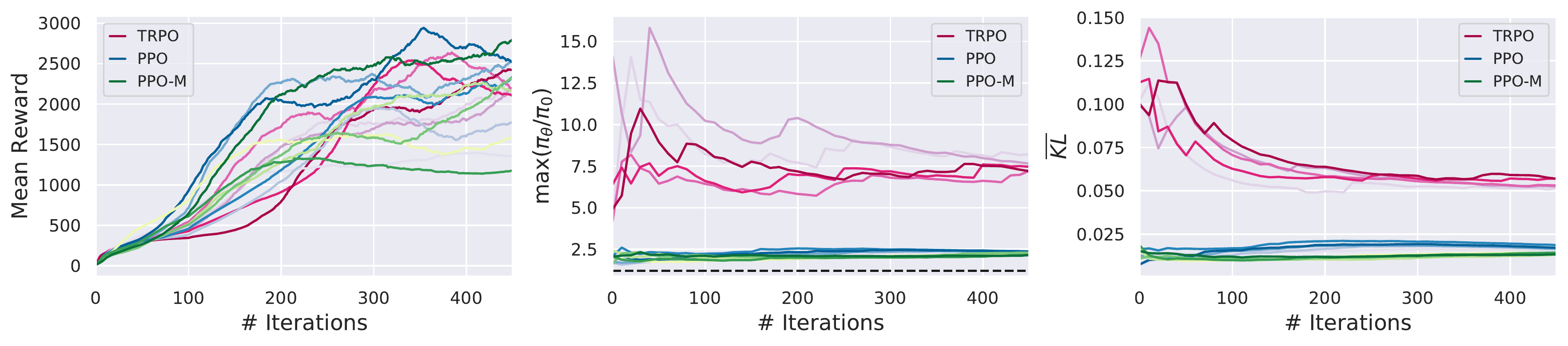} 
		\caption{Hopper-v2 (heldout)}
	\end{subfigure}
	
	\caption{Per step mean reward, maximum ratio (c.f. \eqref{eqn:ppo}), mean KL,
		and maximum versus mean
		KL for agents trained to solve the MuJoCo Humanoid task. The quantities are
		measured over state-action pairs collected from \textit{heldout trajectories}.
		Each line represents a curve from a
		separate agent. The black dotted line
		represents the $1+\epsilon$ ratio constraint in the PPO algorithm, and we measure each
		quantity every twenty five steps. See that the mean KL for TRPO nearly always
		stays within the desired mean KL trust region (at $0.06$).}
	\label{app:trust_regions_heldout}
\end{figure}

%% file: Figures/alg/norm_alg.tex
\begin{algorithm}
\caption{PPO scaling optimization.}\label{alg:ppo-norm}
\begin{algorithmic}[1]
  \Procedure{Initialize-Scaling}{\null}
  \State $R_0 \gets 0$\
  \State $RS = \Call{RunningStatistics}{\null}$ \Comment New running stats class
  that tracks mean, standard deviation 
  \EndProcedure
  \Procedure{Scale-Observation}{$r_t$} \Comment Input: a reward $r_t$
  \State $R_t \gets \gamma R_{t-1} + r_t$\ \Comment $\gamma$ is the reward discount
  \State \Call{Add}{$RS, R_t$}\
  \State \Return $r_t/\Call{Standard-Deviation}{RS}$\ \Comment Returns
  scaled reward
  \EndProcedure
\end{algorithmic}
\end{algorithm}